\documentclass[11pt]{article}
\usepackage[preprint]{acl}
\usepackage{times}
\usepackage{latexsym}
\usepackage{amsmath}
\usepackage{amssymb}
\usepackage{graphicx}
\usepackage{booktabs}
\usepackage{multirow}
\usepackage{float}
\usepackage{tikz}
\usetikzlibrary{arrows.meta, shapes, positioning, backgrounds}
\usepackage[T1]{fontenc}
\usepackage[utf8]{inputenc}
\usepackage{microtype}
\usepackage{inconsolata}
\usepackage{graphicx}

\usepackage{xargs} 
\usepackage[pdftex,usenames,dvipsnames,table,svgnames]{xcolor}
\usepackage[colorinlistoftodos,prependcaption,textsize=tiny]{todonotes} 
\newcommandx{\jz}[2][1=]{\todo[linecolor=Magenta,backgroundcolor=Magenta!25,bordercolor=Magenta,#1]{Jieyu: #2}}

\title{MED-COPILOT: A Medical Assistant Powered by GraphRAG and Similar Patient Case Retrieval}
\author{
Shuheng Chen, Namratha Patil, Haonan Pan, Angel Hsing-Chi Hwang,\\
{\bf Yao Du,  Ruishan Liu, Jieyu Zhao}\\
University of Southern California \\
}

\begin{document}
\maketitle
\begin{abstract}
Clinical decision-making requires synthesizing heterogeneous evidence, including patient histories, clinical guidelines, and trajectories of comparable cases. While large language models (LLMs) offer strong reasoning capabilities, they remain prone to hallucinations and struggle to integrate long, structured medical documents. We present MED-COPILOT, an interactive clinical decision-support system designed for clinicians and medical trainees, which combines guideline-grounded GraphRAG retrieval with hybrid semantic–keyword similar-patient retrieval to support transparent and evidence-aware clinical reasoning. The system builds a structured knowledge graph from WHO and NICE guidelines, applies community-level summarization for efficient retrieval, and maintains a 36,000-case similar-patient database derived from SOAP-normalized MIMIC-IV notes and Synthea-generated records.

We evaluate our framework on clinical note completion and medical question answering, and demonstrate that it consistently outperforms parametric LLM baselines and standard RAG, improving both generation fidelity and clinical reasoning accuracy. The full system is available at \url{https://huggingface.co/spaces/Cryo3978/Med_GraphRAG}
, enabling users to inspect retrieved evidence, visualize token-level similarity contributions, and conduct guided follow-up analysis. Our results demonstrate a practical and interpretable approach to integrating structured guideline knowledge with patient-level analogical evidence for clinical LLMs.

\end{abstract}

\begin{figure}[t]
  \centering
  \includegraphics[width=\linewidth]{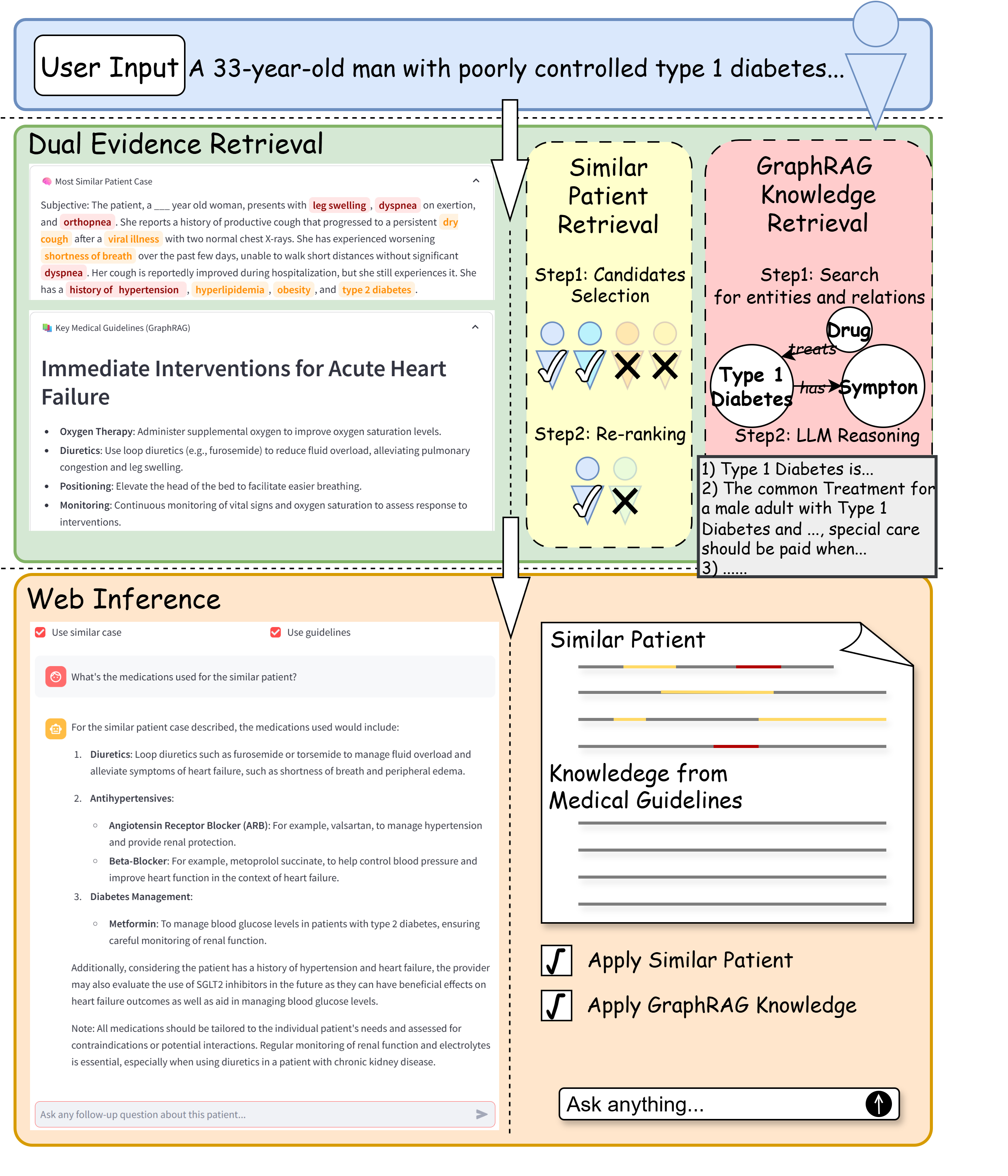}
  \caption{Overview of the proposed system, which performs dual evidence retrieval from similar patient cases and graph-structured clinical guidelines to support clinical inference.}
  \label{fig:workflow}
\end{figure}

\section{Introduction}


Clinical reasoning requires integrating heterogeneous evidence sources, including patient histories, population-level clinical guidelines, and outcomes from comparable prior cases \cite{esteva2019guide, rajkomar2019machine}. While large language models (LLMs) exhibit strong open-domain reasoning capabilities, their use in medical settings remains risky due to hallucinations, confidence miscalibration, and a lack of verifiable evidence grounding \cite{pandit2025medhallu, kim2025limitations, huang2025survey}. Consequently, there is growing consensus that safe and trustworthy medical AI must rely on explicit retrieval of external clinical evidence rather than purely parametric knowledge \cite{nazi2024large, asgari2025framework}, particularly for clinical decision-support systems intended to assist clinicians in real-world reasoning workflows.

Retrieval-Augmented Generation (RAG) frameworks partially address this need by grounding model outputs in retrieved text \citep{lewis2020retrieval, izacard2021leveraging}. However, conventional RAG operates over flat, unstructured corpora and struggles to model the relational and hierarchical structure essential for clinical reasoning \cite{lewis2020retrieval, zhang2025survey}. Recent work has therefore emphasized graph-based retrieval, such as knowledge-graph-augmented RAG and GraphRAG, to support multi-step reasoning over structured clinical knowledge \citep{liang2025kag, sanmartin2024kg, wu-etal-2025-medical}.

Despite these advances, a critical aspect of clinical reasoning remains largely unsupported: patient-level analogical reasoning. In clinical practice, decisions are frequently informed by comparisons to clinically similar patients with shared symptoms, comorbidities, or disease trajectories \cite{beam2018big, rajkomar2018scalable}. Existing clinical RAG systems primarily focus on guideline or document retrieval and rarely incorporate patient-level comparators into the reasoning loop \cite{wu-etal-2025-medical, zhang2025survey, lu2024clinicalrag}, leaving an important gap in case-centric decision.

To bridge these limitations, we propose a unified and interpretable framework that integrates guideline-grounded GraphRAG retrieval, hybrid similar-patient retrieval, and LLM-based reasoning within a single pipeline. The system constructs a structured knowledge graph from authoritative WHO and NICE clinical guidelines \cite{who_guidelines_2024, nice_guidelines_2024} and pairs it with a large-scale similar-patient database derived from SOAP-normalized MIMIC-IV ICU records and Synthea-generated synthetic trajectories \cite{johnson2023mimic, walonoski2018synthea}. A hybrid similarity function combines clinically weighted keyword matching with semantic embedding similarity, enabling flexible and transparent evidence selection during inference.

We evaluate our framework on both generative and discriminative medical reasoning tasks, including clinical note completion on MIMIC-IV and standardized medical question answering benchmarks such as MedQA and the clinical subset of MMLU \cite{jin2021dmlab, hendrycks2021mmlu}. Across settings, the proposed approach consistently outperforms LLM-only and conventional RAG baselines, demonstrating the complementary value of structured guideline knowledge and patient-level analogical evidence. We additionally release an interactive system to support evidence inspection and reproducible clinical reasoning workflows.

Our contributions are threefold: (1) we introduce the first unified framework that jointly incorporates graph-structured guideline retrieval and similar-patient retrieval for clinical reasoning; (2) we construct scalable and standardized clinical evidence resources spanning both guidelines and patient trajectories; and (3) we demonstrate consistent performance gains across multiple medical reasoning tasks with a publicly available interactive system.

\section{Related Work}

\subsection{Retrieval-Augmented Clinical Reasoning}

Retrieval-Augmented Generation (RAG) enhances large language models by grounding generation in externally retrieved evidence, and has become a widely adopted paradigm for medical question answering and clinical reasoning \cite{lewis2020retrieval, izacard2021leveraging, xiong2024benchmarking}. Domain-specific variants adapt RAG to biomedical corpora such as PubMed, clinical guidelines, and electronic health records, demonstrating substantial improvements over LLM-only or chain-of-thought baselines on different benchmarks including MedQA \cite{xiong2024benchmarking, xiong2024improving}.

However, prior studies consistently report that text-based RAG remains sensitive to retrieval noise and degrades under long-context and multi-hop reasoning scenarios, limiting its reliability for complex clinical decision-making \cite{tang2024multihop, sun2025hanrag, gupta2024comprehensive}. To address these limitations, recent work incorporates structured knowledge graphs into retrieval pipelines, enabling explicit modeling of biomedical entities and relations for more interpretable multi-step reasoning \cite{yasunaga2021qa, sohn2025rationale}. GraphRAG-style systems further organize retrieved evidence into graph-structured representations and community-level summaries, improving provenance tracking and reasoning consistency in medical QA and diagnostic tasks \cite{wu-etal-2025-medical, sekar2025investigations, yang2024kg}. Despite these advances, existing systems primarily focus on guideline- or document-level evidence and largely overlook patient-level analogical reasoning.

\subsection{Similar-Patient Retrieval}

Similar-patient retrieval has been extensively studied in intensive care analytics and personalized risk modeling. Early work demonstrates that trajectory-based similarity enables cohort-level outcome estimation and interpretable subgroup analysis in ICU populations \cite{lee2015personalized, alcaide2021visual}. More recent representation-learning approaches model temporal dynamics and relational structures among patients to improve predictive performance \cite{sun2022deep, ma2023predicting}.

Despite its clinical relevance, prior work typically treats patient similarity as a standalone analytical or predictive tool, rather than integrating similar-patient evidence into retrieval-augmented or LLM-based reasoning pipelines \cite{sharafoddini2017patient, zhao2023large}. As a result, patient-level analogical evidence remains underutilized in current clinical RAG systems, motivating the urgent need for unified frameworks that combine structured guideline knowledge with clinically comparable patient trajectories.


\section{Methodology}

This section describes the design of a unified retrieval framework that integrates graph-augmented clinical guidelines and similar-patient evidence. We focus on how authoritative medical knowledge is represented, indexed, and retrieved to support interactive and transparent clinical reasoning.

\subsection{System Overview}

The proposed system supports clinical queries in the form of either SOAP-formatted patient cases or free-text questions. Given an input, it retrieves two complementary forms of evidence: (i) graph-structured guideline knowledge encoding multi-step clinical logic, and (ii) clinically similar patient cases capturing trajectory-level analogical patterns. Retrieved evidence is aggregated and presented with explicit provenance to support downstream reasoning and user inspection.

\subsection{Graph-Augmented Guideline Retrieval}

\paragraph{Guideline Sources.}
To ensure evidential reliability and clinical safety, all guideline knowledge is sourced exclusively from internationally sanctioned authorities. Specifically, we curate clinical practice guidelines from the World Health Organization (WHO) and the National Institute for Health and Care Excellence (NICE), both of which provide consensus-based, evidence-graded recommendations with transparent revision and disclosure procedures \cite{who_guidelines_2024, nice_guidelines_2024}.

After normalization and de-duplication, the resulting corpus comprises 118 WHO guidelines and 525 NICE guidelines, totaling over 40 million tokens of clinically actionable recommendations. These guidelines explicitly articulate eligibility criteria, contraindications, risk boundaries, and escalation pathways, forming a reliable substrate for structured retrieval and multi-hop reasoning.

\paragraph{Graph Construction.}
Clinical guidelines are long, hierarchically structured documents whose logic cannot be fully captured through flat text segmentation alone. While segmenting guidelines into semantically coherent units supports fine-grained retrieval, it fails to expose relational structures such as conditional applicability, treatment dependencies, and escalation logic.

To address this limitation, each guideline is first segmented into semantically coherent \emph{TextUnits}, where each unit corresponds to a single recommendation, decision step, or well-defined clinical scenario. Segmentation respects the native guideline hierarchy rather than arbitrary token windows. Each TextUnit is annotated with metadata derived from the original guideline document, including its source, section title, and character span, enabling transparent provenance tracking and citation.

We then construct a clinical knowledge graph by extracting clinically meaningful entities (e.g., diseases, symptoms, medications, procedures) and relations (e.g., indication, contraindication, monitoring, escalation) from the segmented TextUnits. Entity mentions are identified using medical named-entity recognition and domain-adapted LLM prompting, and normalized to standard biomedical vocabularies (e.g., SNOMED-CT, ICD-10, ATC). Relations and constraint qualifiers (such as age limits or escalation triggers) are encoded explicitly, and all graph elements retain links to their originating guideline sections. This graph representation enables structured navigation of guideline logic beyond only what is accessible through text-only retrieval.

\paragraph{Indexing.}
To support efficient retrieval across both fine-grained recommendations and higher-level guideline logic, we adopt a hybrid indexing strategy. Guideline TextUnits, graph community summaries, and entity descriptions are embedded using BioClinicalBERT and stored in a local vector database (LanceDB). Indexing is organized by evidence modality, preserving the structural hierarchy of the guideline graph while enabling flexible retrieval at different levels of granularity with full provenance traceability.

\subsection{Similar-Patient Retrieval}

\paragraph{Patient Data Sources.}
Similar-patient retrieval operates over a curated repository of 36{,}000 structured patient cases. To balance clinical realism and controlled coverage, the repository integrates two complementary sources. First, 18{,}000 intensive care encounters are sampled from MIMIC-IV, a large-scale de-identified critical care database containing comprehensive EHR-derived clinical documentation \cite{johnson2023mimic}. Each admission is converted into a standardized SOAP representation using a rule-guided conversion agent, preserving temporal evolution and diagnostic intent while normalizing stylistic variation.

Second, 18{,}000 additional patient trajectories are generated using Synthea, an agent-based synthetic patient simulator designed to model longitudinal disease courses and care pathways \cite{walonoski2018synthea}. These synthetic records are rewritten under the same SOAP specification to align structure and semantic granularity with the MIMIC-derived cohort. The combined repository thus maintains structural uniformity while retaining clinically meaningful diversity in etiology, acuity, and disease progression.

\paragraph{Hybrid Similarity Scoring.}
Identifying clinically similar patients requires balancing strict clinical constraints with trajectory-level similarity, motivating a hybrid retrieval formulation that combines keyword alignment with semantic embeddings.

To identify clinically comparable cases, we rank candidate patients using a hybrid similarity formulation that combines keyword-conditioned alignment with embedding-based semantic proximity. The keyword component emphasizes discrete clinical signals such as diagnoses, comorbidities, and key interventions, while the semantic component captures graded similarity in longitudinal trajectories and response patterns. This dual formulation allows retrieval to satisfy eligibility-level constraints while remaining robust to surface-level lexical variation. For conciseness, we present only the high-level formulation here; detailed mathematical definitions are provided in the appendix.

\subsection{Unified Retrieval Workflow}

At inference time, a query case or question is embedded and used to retrieve semantically relevant guideline artifacts and patient cases from the vector index. In parallel, graph-conditioned retrieval identifies guideline communities whose relational structure aligns with the query, enabling access to multi-hop clinical logic such as contraindications, escalation pathways, and applicability constraints.

Retrieved guideline evidence and similar patient cases are then aggregated into a unified evidence set comprising ranked guideline communities with supporting nodes and ranked patient cases with aligned attributes. All retrieved items are presented with explicit source links, allowing users to inspect provenance and understand the basis of the system’s recommendations.

Details of the GraphRAG construction and retrieval pipeline
are provided in Appendix~B.

\section{Evaluation and Demonstration}

We evaluate the proposed framework on clinical note generation and medical question answering (QA), and additionally present an interactive system demonstration.

\subsection{Tasks and Setup}

\paragraph{Clinical Note Generation.}
We sample 1{,}000 de-identified ICU notes from MIMIC-IV and convert each into standardized SOAP format \cite{weed1968soap}. The model receives the Subjective (S), Objective (O), and Assessment (A) sections and generates the corresponding Plan (P).

\paragraph{Medical QA.}
We evaluate on MedQA \cite{jin2021dmlab} and the clinical subset of MMLU \cite{hendrycks2021mmlu} using the same retrieval-augmented pipeline as in the clinical setting.

The similarity mixing weight \( \lambda \) is tuned on a 100-case validation set and fixed across all experiments.

\subsection{Metrics and Baselines}

For note generation, we report ROUGE-L and BERTScore-F\textsubscript{1}; additional metrics are provided in the appendix.  
For QA, we report standard accuracy.

We compare against strong parametric-only LLM baselines (DeepSeek-Chat, GPT-4.1-mini, and Gemini~2.5) under identical prompting and decoding settings.

\subsection{Main Results}

Table~\ref{tab:main_results} summarizes performance across tasks. Retrieval augmentation consistently improves both generation and QA performance. Combining similar-patient retrieval and GraphRAG yields the strongest results across all backbone models.

\begin{table*}[t]
\centering
\small
\begin{tabular}{lccccccccc}
\toprule
\multirow{3}{*}{\textbf{Model}} 
  & \multicolumn{6}{c}{\textbf{Text Generation}} 
  & \multicolumn{2}{c}{\textbf{QA}} \\
\cmidrule(lr){2-7} \cmidrule(lr){8-9}
  & \multicolumn{6}{c}{\textbf{MIMIC-IV Clinical Notes}} 
  & \textbf{MedQA} & \textbf{MMLU-clinical} \\
\cmidrule(lr){2-7} \cmidrule(lr){8-9}
  & BLEU & METEOR & ROUGE-1 & ROUGE-2 & ROUGE-L & BERTScore-F1 
  & Acc & Acc \\
\midrule
\multicolumn{9}{c}{\textbf{Baselines without Retrieval}} \\
\midrule
DeepSeek-Chat  & 0.006 & 0.251 & 0.205 & 0.042 & 0.120 & 0.809 & 0.809 & 0.857 \\
GPT-4.1-mini   & 0.012 & 0.282 & 0.301 & 0.064 & 0.144 & 0.825 & 0.846 & 0.868 \\
Gemini~2.5     & 0.014 & 0.289 & 0.295 & 0.070 & 0.147 & 0.829 & 0.867 & 0.875 \\
\midrule
\multicolumn{9}{c}{\textbf{with RAG}} \\
\midrule
DeepSeek-Chat  & 0.015 & 0.272 & 0.229 & 0.057 & 0.157 & 0.799 & 0.851 & 0.875 \\
GPT-4.1-mini   & 0.019 & 0.291 & 0.351 & 0.081 & 0.221 & 0.831 & 0.885 & 0.887 \\
Gemini~2.5     & 0.020 & 0.316 & 0.368 & 0.099 & 0.214 & 0.833 & 0.892 & 0.905 \\
\midrule
\multicolumn{9}{c}{\textbf{Our Framework}} \\
\midrule
DeepSeek-Chat  & 0.029 & 0.287 & 0.340 & 0.097 & 0.203 & 0.839 & 0.864 & 0.910 \\
GPT-4.1-mini   & 0.035 & 0.321 & 0.380 & 0.124 & 0.291 & 0.851 & 0.912 & 0.971 \\
Gemini~2.5     & \textbf{0.042} & \textbf{0.334} & \textbf{0.419} & \textbf{0.155} & \textbf{0.300} & \textbf{0.862} & \textbf{0.937} & \textbf{0.971} \\
\bottomrule
\end{tabular}
\caption{Performance under different retrieval configurations on clinical note generation and QA benchmarks.}
\label{tab:main_results}
\end{table*}

\subsection{Ablation}

Table~\ref{tab:ablation} isolates the contribution of each retrieval component. Both similar-patient retrieval and GraphRAG knowledge retrieval independently improve performance, with the combined framework achieving the largest gains.

\begin{table*}[t]
\centering
\small
\begin{tabular}{lccccccccc}
\toprule
\multirow{3}{*}{\textbf{Setting}}
  & \multicolumn{6}{c}{\textbf{Text Generation}} 
  & \multicolumn{2}{c}{\textbf{QA}} \\
\cmidrule(lr){2-7} \cmidrule(lr){8-9}
  & \multicolumn{6}{c}{\textbf{MIMIC-IV Clinical Notes}} 
  & \textbf{MedQA} & \textbf{MMLU-clinical} \\
\cmidrule(lr){2-7} \cmidrule(lr){8-9}
  & BLEU & METEOR & ROUGE-1 & ROUGE-2 & ROUGE-L & BERTScore-F1 
  & Acc & Acc \\
\midrule
Baseline      & 0.012 & 0.282 & 0.301 & 0.064 & 0.144 & 0.825 & 0.846 & 0.868 \\
+ SPR         & 0.028 & 0.315 & 0.324 & 0.099 & 0.159 & 0.828 & 0.851 & 0.914 \\
+ GraphRAG    & 0.031 & 0.310 & 0.352 & 0.115 & 0.258 & 0.840 & 0.870 & 0.952 \\
+ Both        & \textbf{0.035} & \textbf{0.321} & \textbf{0.380} & \textbf{0.124} & \textbf{0.291} & \textbf{0.851} & \textbf{0.912} & \textbf{0.971} \\
\bottomrule
\end{tabular}
\caption{Ablation results isolating the contribution of similar-patient retrieval (SPR) and GraphRAG-based guideline reasoning.}
\label{tab:ablation}
\end{table*}

\subsection{Interactive System}

We deploy the proposed framework as a public HuggingFace Space.\footnote{\url{https://huggingface.co/spaces/Cryo3978/Med_GraphRAG}} 
The interface provides end-to-end interactive clinical exploration grounded in dual evidence retrieval, enabling clinicians to inspect, compare, and reason over guideline-level and patient-level evidence within a unified decision-support workflow.

\paragraph{Dual Evidence Control.}
Given a free-text clinical case, the system retrieves (i) similar patient records from the case repository and (ii) guideline-level evidence via the GraphRAG pipeline. 
Users may selectively enable or disable each evidence source through explicit interface controls, allowing controllable reasoning under patient-level evidence, guideline-level knowledge, or their combination. 
The selected evidence is automatically packaged with the query context for downstream generation and question answering.

\paragraph{Query-Conditioned Saliency Highlighting.}
To improve interpretability during evidence inspection, the similar-patient panel incorporates a lightweight query-conditioned saliency overlay.
Each extracted clinical concept (e.g., symptoms, diagnoses, comorbidities) is assigned an importance score based on its semantic similarity to the input query embedding.

The score is discretized into two levels and visually encoded (yellow: important; red: highly important), enabling users to quickly identify case attributes most relevant to the current clinical question.

\begin{figure}[H]
    \centering
    \includegraphics[width=\linewidth]{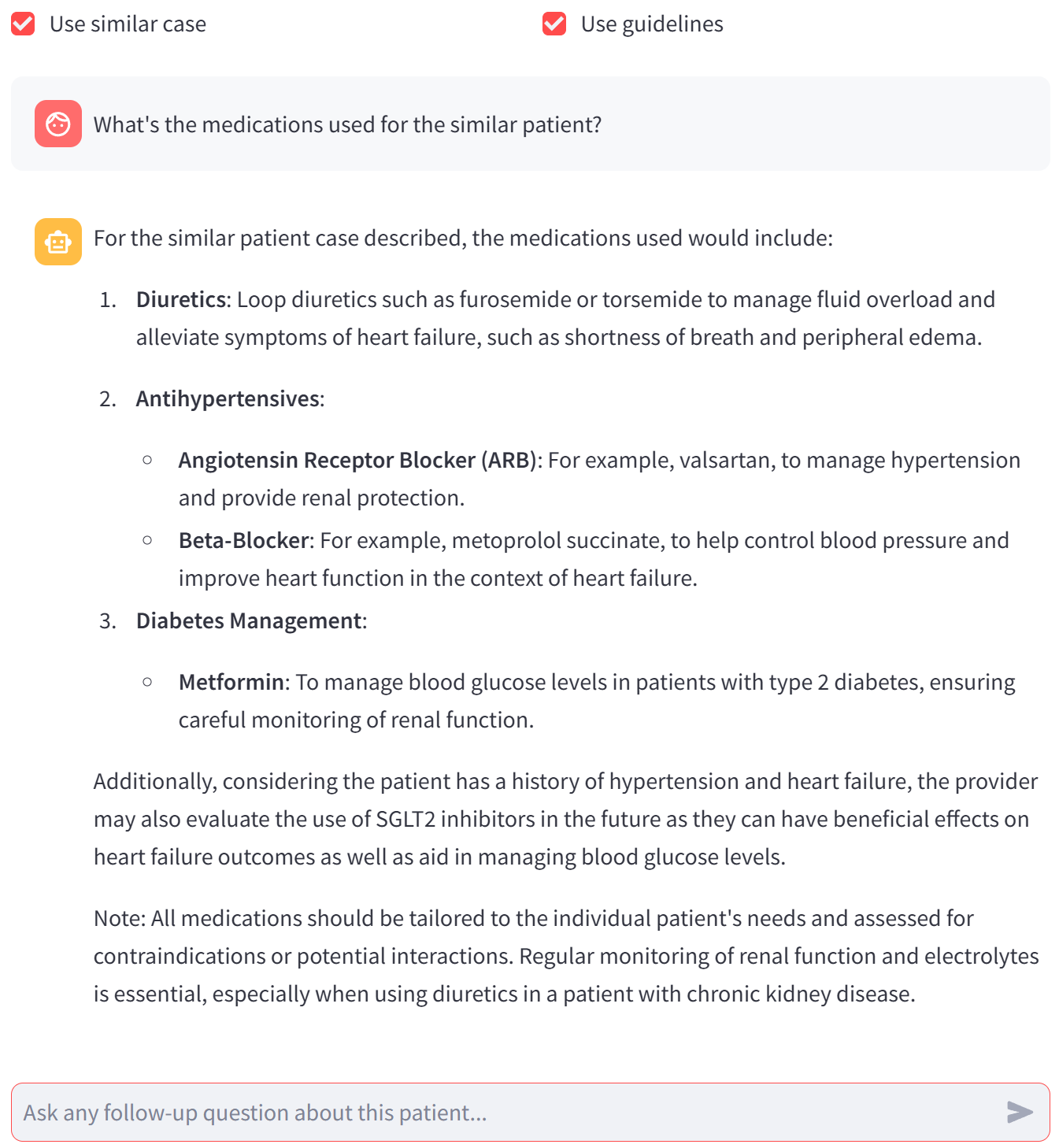}
    \caption{Interactive question answering with dual evidence control.}
    \label{fig:interactive_qa}
\end{figure}

\begin{figure}[H]
    \centering
    \includegraphics[width=\linewidth]{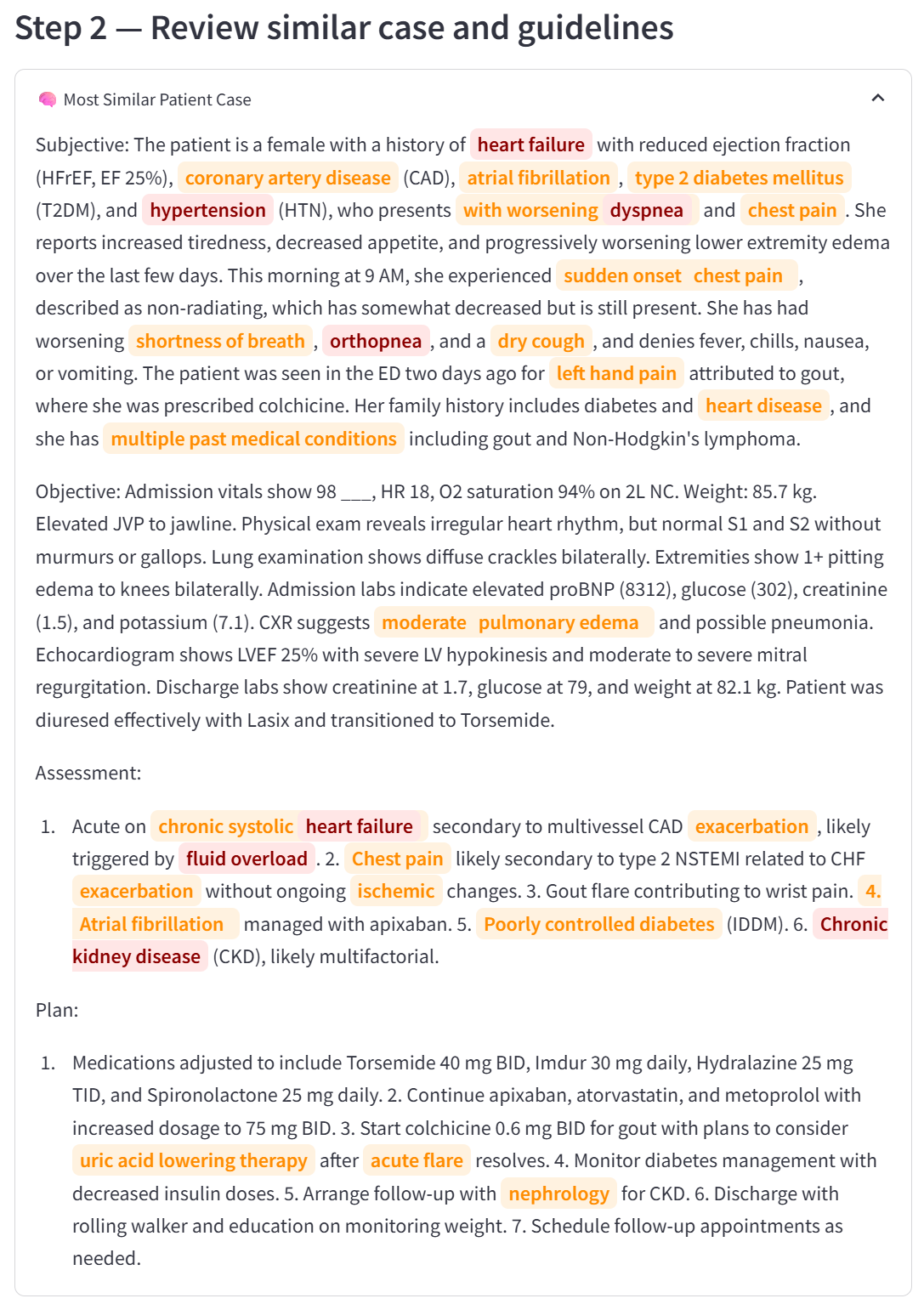}
    \caption{Similar-patient panel with query-conditioned saliency highlighting. 
    Clinical concepts are color-coded according to their semantic relevance to the input query (yellow: important; red: highly important).}
    \label{fig:saliency_view}
\end{figure}

Figure~\ref{fig:interactive_qa} illustrates interactive question answering with controllable evidence selection. 
Figure~\ref{fig:saliency_view} shows the similar-patient view with query-conditioned saliency highlighting.

Additional screenshots and a step-by-step walkthrough are provided in Appendix~A.

\section{Limitations}

The proposed system has several limitations. First, the similar-patient
repository combines SOAP-normalized MIMIC-IV records with Synthea-generated
synthetic cases, resulting in heterogeneous data distributions that only
partially reflect real-world clinical documentation and temporal complexity.
Second, the guideline knowledge graph abstracts narrative recommendations into
structured relations, which may omit contextual nuances present in full
guideline texts. Third, the system is a research prototype intended for
decision-support and evidence inspection rather than autonomous clinical use.
Finally, evaluation is limited to note completion and medical QA benchmarks,
and no gold-standard dataset currently exists for assessing patient-level
similarity retrieval. 

Future work will explore real-world clinical deployment settings, expanded
guideline coverage, and uncertainty-aware reasoning mechanisms to further
improve safety, robustness, and practical applicability.

\section{Conclusion}

We presented MED-COPILOT, an interactive clinical decision-support system
designed to assist clinicians and trainees in transparent and evidence-aware reasoning.
By integrating guideline-grounded GraphRAG retrieval with similar-patient
evidence within a unified and interpretable pipeline, the system enables
clinicians to inspect, compare, and reason over multi-source clinical
evidence. Experimental results demonstrate that combining structured
guideline knowledge with patient-level analogical retrieval provides
complementary benefits across clinical note generation and medical QA tasks,
highlighting the value of controllable and inspectable evidence integration
for clinical AI systems. Future work will focus on expanding guideline coverage, improving uncertainty-aware reasoning, and exploring deployment in real-world clinical workflows.

\section*{Ethical Statement}

This work uses the MIMIC-IV database under its approved data-use agreement, and
all authors completed the required credentialing. Synthetic cases generated
with Synthea do not correspond to real patients. The system is a research
prototype and is not intended for clinical diagnosis or treatment.

\bibliography{custom}

\newpage
\appendix
\section{Demo Walkthrough}
\label{sec:demo_walkthrough}

This appendix provides a step-by-step walkthrough of the interactive demo
interface.

\subsection{Case Input and Query Locking}

\begin{figure}[H]
    \centering
    \includegraphics[width=\linewidth]{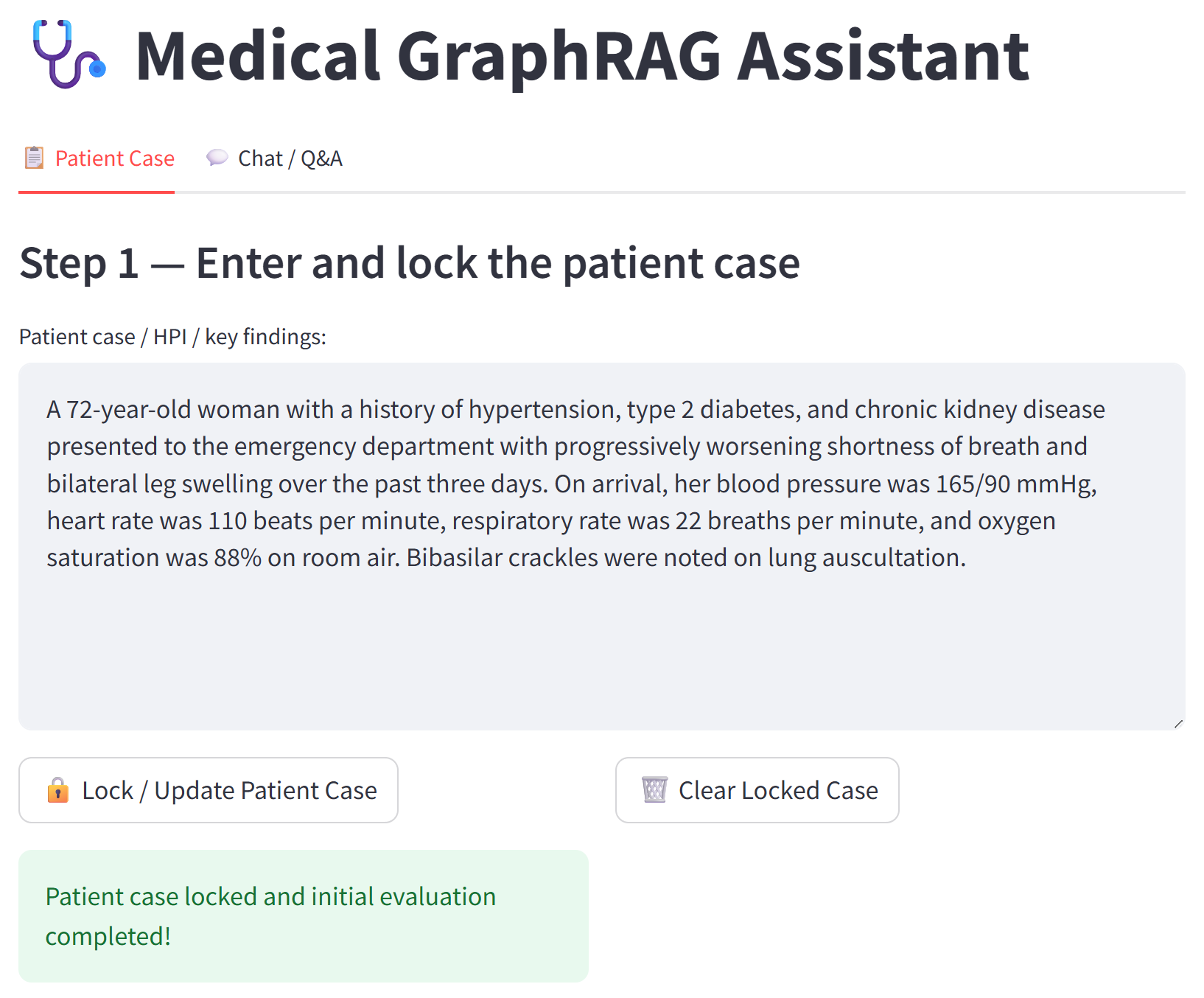}
    \caption{Patient case input and query locking interface.}
    \label{fig:queryinput}
\end{figure}


\subsection{Similar Patient Retrieval}

\begin{figure}[H]
    \centering
    \includegraphics[width=\linewidth]{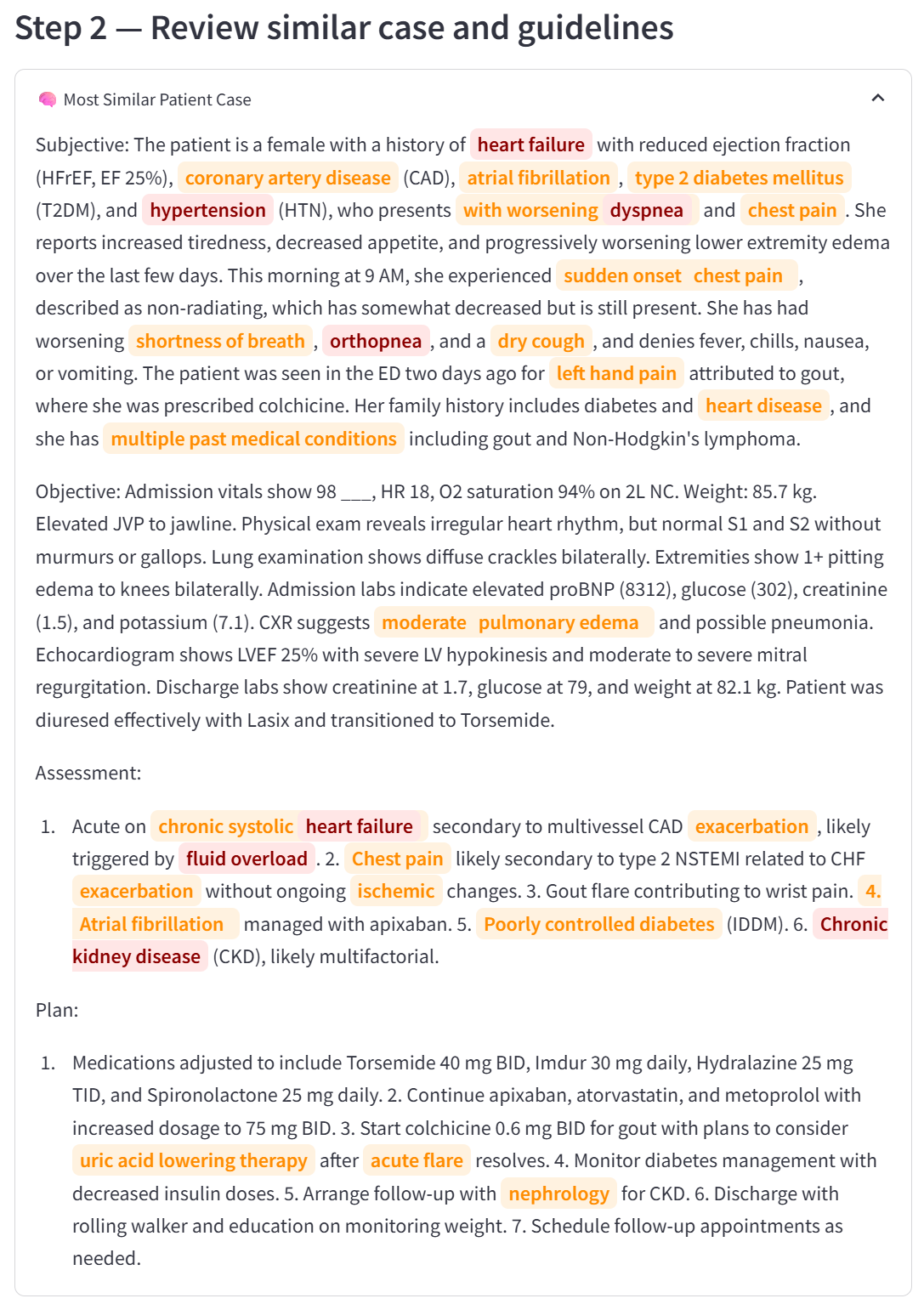}
    \caption{Retrieval of the most similar patient case from the curated
    repository.}
    \label{fig:similarpatient}
\end{figure}

Following case locking, the system retrieves clinically similar patient
records from the 36{,}000-case repository using the proposed hybrid similarity
scoring function. Figure~\ref{fig:similarpatient} shows the most similar patient
identified for the input case. Clinically salient attributes, including
symptoms, comorbidities, diagnostic findings, and interventions, are
highlighted to support rapid comparison and interpretability. This retrieved
case provides experiential, case-based context grounded in prior patient
trajectories.

\subsection{Guideline Retrieval via GraphRAG}

\begin{figure}[H]
    \centering
    \includegraphics[width=\linewidth]{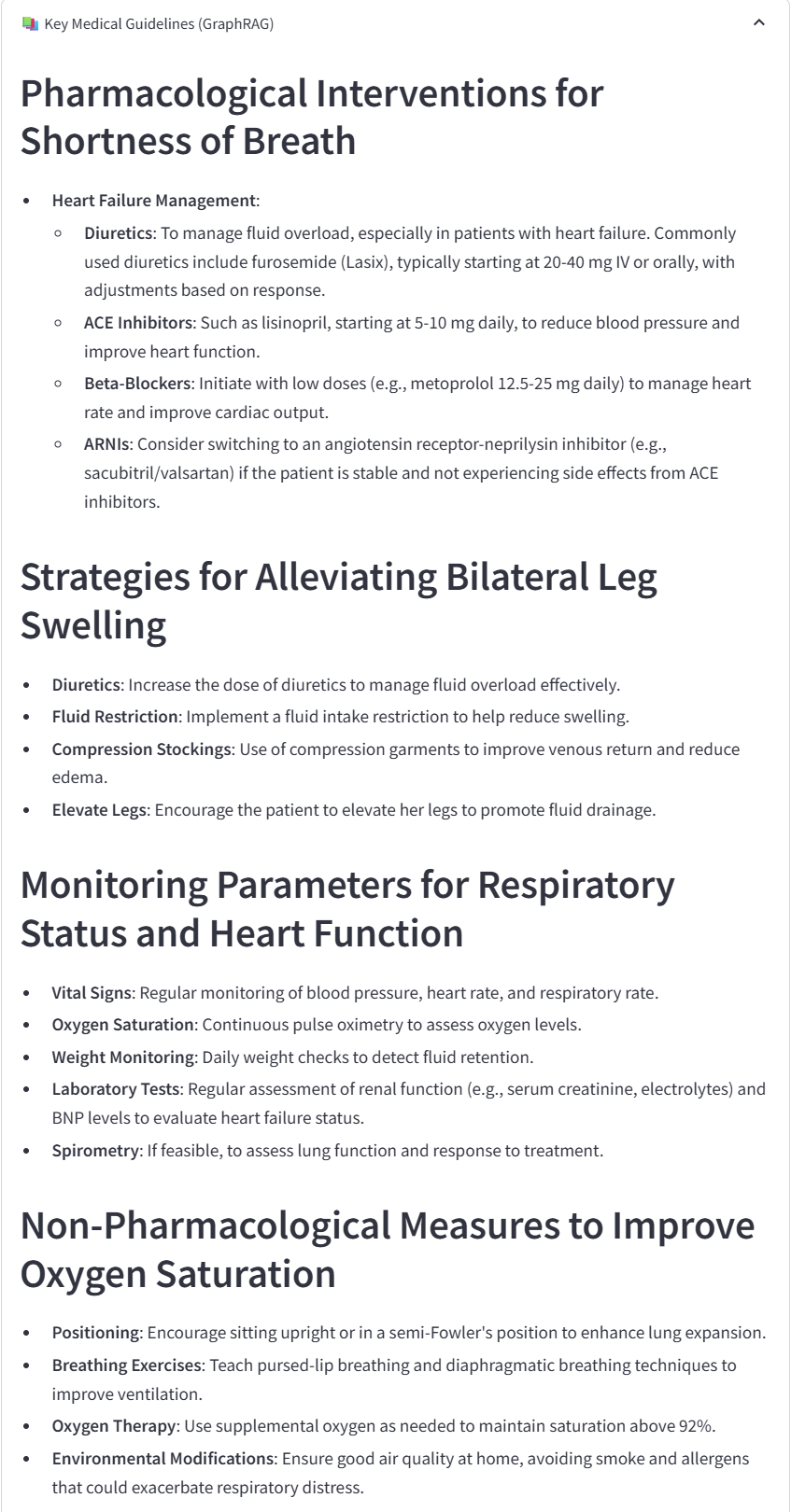}
    \caption{Guideline-grounded evidence retrieval through the GraphRAG
    pipeline.}
    \label{fig:GraphRAGknowledge}
\end{figure}

In parallel with similar-patient retrieval, the system performs
guideline-grounded retrieval using the GraphRAG pipeline. As illustrated in
Figure~\ref{fig:GraphRAGknowledge}, relevant clinical recommendations are identified
and summarized from authoritative WHO and NICE guidelines. Retrieved guideline
content exposes structured evidence such as treatment options,
contraindications, and escalation criteria.

\subsection{Interactive Evidence-Aware Question Answering}

\begin{figure}[H]
    \centering
    \includegraphics[width=\linewidth]{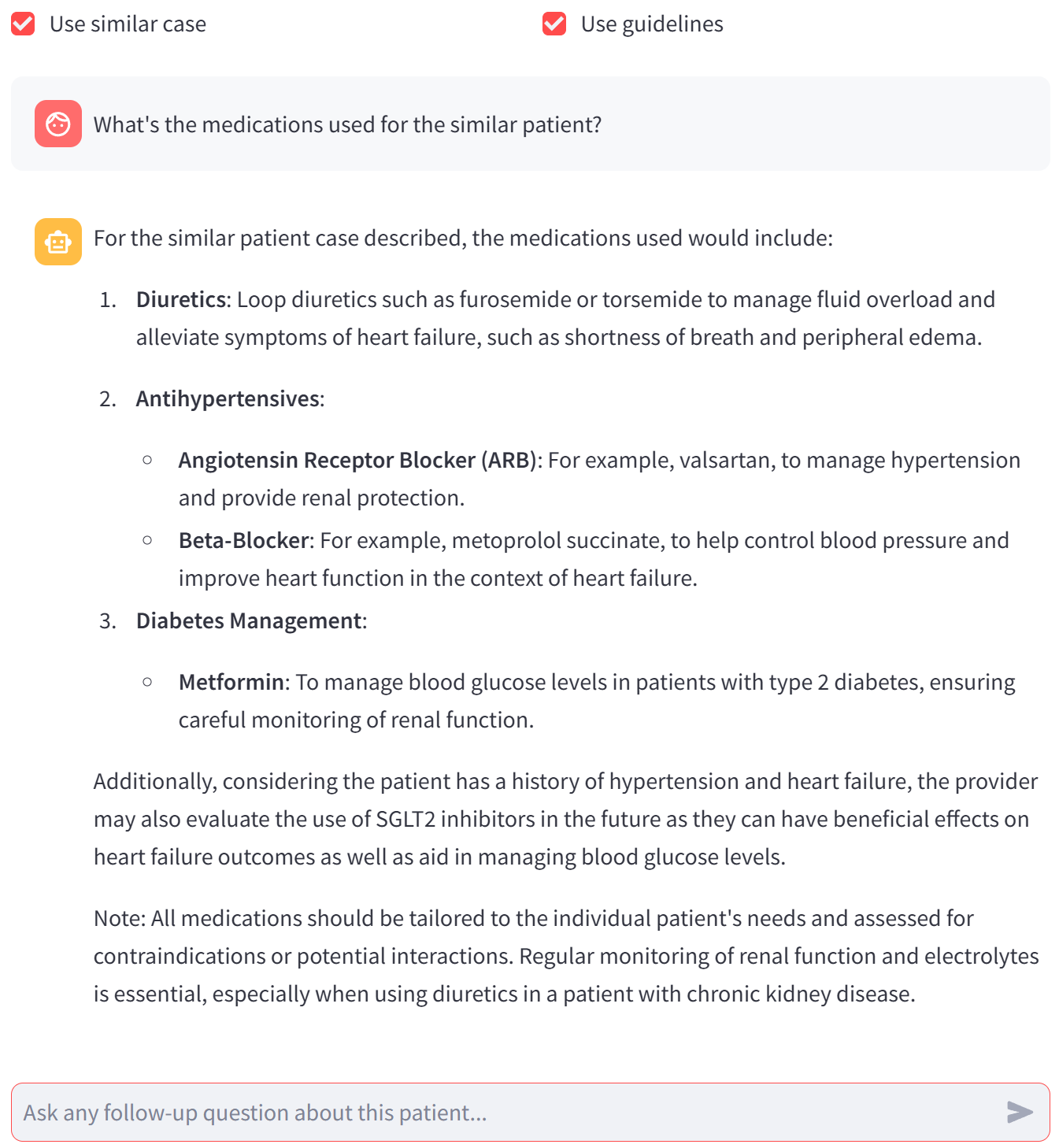}
    \caption{Interactive question answering with user-controlled evidence
    selection.}
    \label{fig:demopage4}
\end{figure}

Figure~\ref{fig:demopage4} demonstrates the interactive question-answering
interface. Users may selectively choose whether to incorporate similar patient
cases, guideline summaries, or both into downstream reasoning. Based on the
selected evidence sources, the system dynamically constructs a structured
prompt combining the locked patient case and retrieved evidence, enabling
transparent and controllable clinical exploration. Generated responses
explicitly reflect the chosen evidence context, allowing users to examine how
different information sources influence the model’s reasoning.

\section{Appendix B: GraphRAG Construction Details}

\begin{figure}[H]
    \centering
    \includegraphics[width=\linewidth]{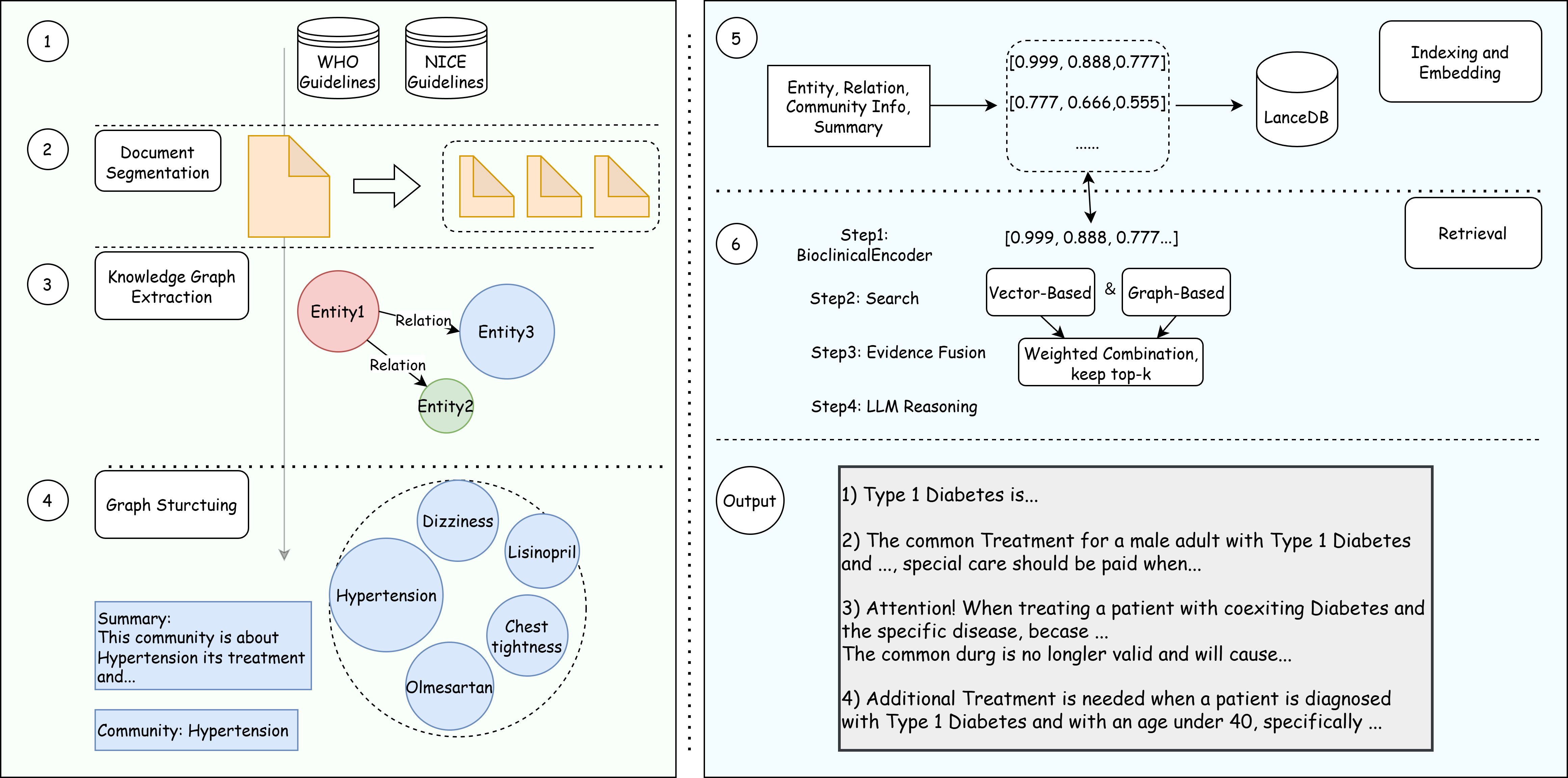}
    \caption{Detailed construction and retrieval pipeline of the GraphRAG module,
    including segmentation, graph extraction, indexing, and hybrid retrieval.}
    \label{fig:graphrag_appendix}
\end{figure}

\end{document}